\documentclass{article}
\usepackage{graphicx}

\PassOptionsToPackage{numbers, sort&compress}{natbib}

\usepackage{hyperref}
\usepackage[preprint]{neurips_2025}
\usepackage{xcolor}
\usepackage{amssymb}
\usepackage{amsmath}
\usepackage{enumitem}
\usepackage{algorithm}
\usepackage{algpseudocode}
\usepackage{placeins} 
\usepackage{xspace}
\usepackage{soul}
\usepackage{wrapfig}
\usepackage{subcaption}

\definecolor{myred}{rgb}{1.0,0.7,0.8}
\definecolor{mygreen}{RGB}{0,166,0}
\definecolor{lightgreen}{rgb}{0.56, 0.93, 0.56}
\definecolor{myorange}{RGB}{252,107,4}
\definecolor{darkgreen}{RGB}{0,153,102}
\definecolor{lightblue}{rgb}{0.53, 0.81, 0.92}
\definecolor{lightgray}{gray}{0.9}

\newtheorem{theorem}{Theorem}

\newtheorem{lemma}[theorem]{Lemma}

\let\svthefootnote\thefootnote
\newcommand\marklessfootnote[1]{%
  \let\thefootnote\relax%
  \footnotetext{#1}%
  \let\thefootnote\svthefootnote%
}

\newcommand{\name}{\texttt{SART}\xspace}

\title{Thinking Short and Right Over Thinking Long: Serving LLM Reasoning Efficiently and Accurately}

\author{
{Yuhang Wang} \And {Youhe Jiang} \And {Bin Cui} \And {Fangcheng Fu}
}

\begin{document}
\maketitle

\begin{abstract}
Recent advances in test-time scaling suggest that Large Language Models (LLMs) can gain better capabilities by generating Chain-of-Thought reasoning (analogous to human thinking) to respond a given request, and meanwhile exploring more reasoning branches (i.e., generating multiple responses and ensembling them) can improve the final output quality.
However, when incorporating the two scaling dimensions, we find that the system efficiency is dampened significantly for two reasons.
Firstly, the time cost to generate the final output increases substantially as many reasoning branches would be trapped in the over-thinking dilemma, producing excessively long responses.
Secondly, generating multiple reasoning branches for each request increases memory consumption, which is unsuitable for LLM serving since we can only batch a limited number of requests to process simultaneously.
To address this, we present \name, a serving framework for efficient and accurate LLM reasoning. 
The essential idea is to manage the thinking to be \textit{short and right, rather than long}.
For one thing, we devise a redundant sampling with early stopping approach based on empirical observations and theoretic analysis, which increases the likelihood of obtaining \textit{short-thinking} responses when sampling reasoning branches. 
For another, we propose to dynamically prune low-quality branches so that only \textit{right-thinking} branches are maintained, reducing the memory consumption and allowing us to batch more requests. 
Experimental results demonstrate that \name not only improves the accuracy of LLM reasoning but also enhances the serving efficiency, outperforming existing methods by up to 28.2$\times$ and on average 15.7$\times$ in terms of efficiency when achieving the same level of accuracy. 
\end{abstract}

\marklessfootnote{Correspondence to: Fangcheng Fu (ccchengff@gmail.com).}

\section{Introduction}
\label{sec:intro}

Large Language Models (LLMs) have demonstrated astonishing effectiveness in various domains like machine translation, document summarization, and dialogue systems \cite{attn,chatgpt,gpt4,zhu2023multilingual,kocmi2023large,iyer2023towards}. 
The growing demand of leveraging LLMs in more real-world applications, in turn, has pushed extensive research into enhancing LLMs' capabilities to solve complex problems. 
Particularly, test-time scaling techniques, which increase the amount of computation during inference to enhance the performance of LLMs, have recently created a series of breakthroughs \cite{snell2024scaling,zhang2025and}.

There are two primary dimensions of test-time scaling. 
The first dimension is sequential scaling with Chain-of-Thought (CoT) reasoning, which drives the LLMs to engage in CoT-like reasoning processes to refine their response. 
Notable examples include OpenAI o1 \cite{jaech2024openai}, DeepSeek R1 \cite{guo2025deepseek}, and Kimi k1.5 \cite{team2025kimi}. 
The second is parallel scaling by branch sampling, which leverages the same model to generate multiple candidate outputs and then ensembles these results to deliver a final response through strategies like majority voting or reward model ranking \cite{wang2022self,aggarwal2023let,irvine2023rewarding,chen2024more}.

\begin{figure}[!t]
\centering
\includegraphics[width=\textwidth]{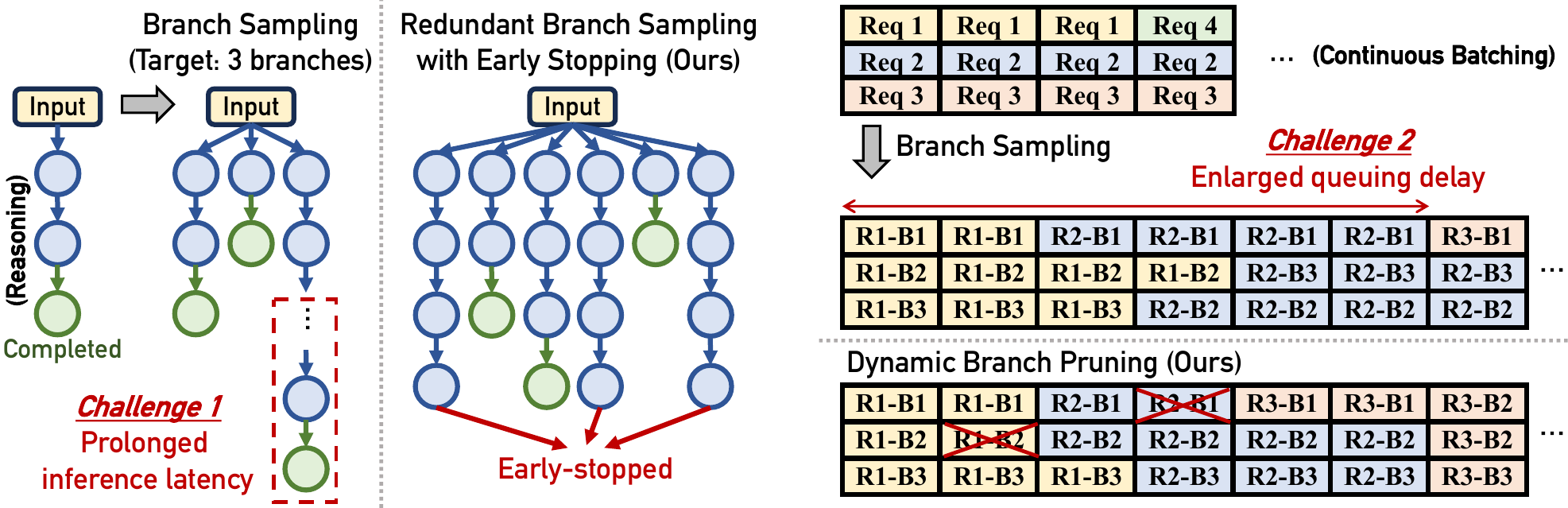}
\caption{\small{Illustration of the two key challenges and our solutions.}}
\label{fig:challenges}
\end{figure}

Although both scaling dimensions have shown remarkable successes in enhancing the quality of LLM responses, integrating them poses significant challenges to the system efficiency. 
We illustrate two key challenges in Figure~\ref{fig:challenges} and elaborate below.
\begin{itemize}[leftmargin=*]
\item \textbf{Single-request inference latency prolonged to the longest branch:}
As state-of-the-art LLMs rely on CoT-like reasoning to generate a response, many studies (as well as our empirical testbed in Section~\ref{sec:motivation}) have shown that different reasoning trials vary in length substantially.
When using branch sampling, although different branches can be batched together for concurrent processing, the inference latency of each request is strongly related to the branch that generates the longest response. 
Worst still, to collect more responses for the final decision, it is more likely that certain branches may engage in extremely long reasoning, prohibitively increasing the inference latency.
\item \textbf{Enlarged queuing delay when serving multiple requests:}
Current works on branch sampling have only considered the efficiency of LLM reasoning for merely a single inference request. 
However, in practice, branch sampling raises hurdles for serving multiple requests. 
In particular, it is well known that efficient LLM serving necessitates batching multiple requests together due to the memory-bounded nature of the decoding phase~\cite{yu2022orca,patel2024splitwise}. 
If we allow for a large number of branches for each request, it would saturate the hardware resources (e.g., computation and memory) easily, so the number of requests that can be processed together would be limited. Consequently, it significantly increases the queuing latency of most requests.
\end{itemize}
To fill this gap, we introduce a brand new framework, namely \name, to serve LLM reasoning efficiently and accurately. The major contributions of this work are as follows.

Firstly, we empirically find that the correctness of responses is weakly related to response lengths. Following this, we design an approach of redundant sampling with early stopping. 
As shown in Figure~\ref{fig:challenges} (left), we begin by sampling more branches, but terminate the remaining ones once a sufficient number of branches have completed. 
Though being simple, we prove that such an approach effectively increases the likelihood of obtaining \textit{short-thinking} responses, and therefore avoids being prolonged by excessively long-thinking branches. 

Secondly, we inspect the resource consumption in branch sampling and observe that many branches consume resources for a long time, even though they are insignificant to the final response. 
To tackle this, we propose a two-phase dynamic pruning method, which judges the quality of branches via a process reward model and prunes low-quality branches in an exploration-exploitation manner. 
By doing so, we prioritize \textit{right-thinking} branches and release the resources consumed by the others on the fly, so that we can reduce the queuing latency effectively, as shown in Figure~\ref{fig:challenges} (right). 

Subsequently, we develop \name, a serving framework for efficient and accurate LLM reasoning. 
\name is equipped with a scheduling strategy that integrates the two techniques with continuous batching, managing the LLM reasoning processes to be short and right for efficiency and accuracy.

Finally, experimental results show that \name improves both the serving efficiency and response accuracy at the same time, and outperforms existing works by up to 28.2$\times$ and on average 15.7$\times$ in terms of efficiency when achieving a comparable level of accuracy.

\section{Background and Related Works}
\label{sec:bg}

\textbf{LLM inference and serving.}
Given an input request (a.k.a. prompt), the LLM inference process consists of two phases: the computation-intensive prefilling phase that processes the entire input and generates the first token in one step, and the memory-bounded decoding phase that generates the subsequent tokens in multiple steps, with each step generating only one token. 
In LLM serving, due to the memory-bounded nature of the decoding phase, it is common practice to batch multiple requests together to improve GPU utilization~\cite{yu2022orca,patel2024splitwise}. 
Since requests arrive at different time, continuous batching, which allows dynamically excluding completed/terminated requests from the current batch and including new requests, plays a critical role.
When the current batch cannot accommodate all pending requests, some of them must wait to be scheduled, so the end-to-end latency of a request includes queuing latency (the time spent waiting for batch availability) and inference latency (the time for prefilling and decoding).

\textbf{LLM reasoning.}
The reasoning capability of LLMs can be largely attributed to the sequential scaling during inference. 
To be specific, inspired by humans’ thinking process, Wei et al.~\cite{wei2023chainofthoughtpromptingelicitsreasoning} found that by generating relatively long CoT-like reasoning before giving the final answer, LLMs become more capable of solving complex problems like mathematic/scientific reasoning and multi-hop question answering.
Driven by this, the training of state-of-the-art LLMs typically involves a reinforcement learning process, such as PPO~\cite{schulman2017proximal_ppo} and GRPO~\cite{shao2024deepseekmath_grpo}, to gain reasoning capabilities. 
It is noteworthy that although the reinforcement learning process does not explicitly control the output length, there is a trend that the trained model gradually learns to give longer outputs to facilitate deeper reasoning~\cite{guo2025deepseek,team2025kimi}. 
Furthermore, many works have reported that reasoning-capable LLMs occasionally get trapped in the over-thinking dilemma, where the model repeatedly switches or even turns over the reasoning process, resulting in extremely long responses~\cite{yi2025shorterbetter,su2025between,chen2025towards,sui2025stop}.

\textbf{Branch sampling.} 
Parallel scaling leverages stochastic sampling to generate diverse candidate outputs from the same model, and then aggregates these outputs to produce a consensus or return the optimal response. 
Typically, multiple reasoning branches (i.e., the CoT-like reasoning trajectories) are generated independently, and the final answer is selected via specific strategies like majority voting or reward ranking~\cite{wang2022self,aggarwal2023let,irvine2023rewarding,chen2024more}. 
For instance, Self-Consistency~\cite{wang2022self} samples a number of branches individually and selects the most frequent answer, reaching a consensus to mitigate errors of individual reasoning trajectories, while Best-of-$N$~\cite{irvine2023rewarding} also generates individual branches and ranks them by domain-specific metrics or some external reward model. 
There are also works that integrate sequential and parallel scaling via complex tree-based searching algorithms~\cite{browne2012survey_mcts,lin2025leveraging,yao2023tree,wu2024inference}. For instance, Rebase~\cite{wu2024inference} constructs a tree of reasoning trajectories, and iteratively expands a node (deepening the reasoning sequentially) or samples more branches (broadening the reasoning in parallel) under the guidance of a reward model.

Given the long output length, the inference latency of LLM reasoning is primarily dominated by the decoding phase, with longer reasoning requiring more decoding steps and thereby higher inference latency.
When incorporating branch sampling, since different branches are generated individually, we can batch them together during the decoding phase, making the inference latency related to the longest branch. 
However, as we will show in Section~\ref{sec:motivation}, even for the same request, different trials of LLM reasoning would greatly vary in length. 
To generate multiple reasoning trajectories, it is more frequent that the model exhibits over-thinking in some branches, which slows down the inference process to a large extent. 
In addition, current works mainly focus on enhancing the response quality for a single request, while how to serve multiple requests is underexplored. 
To be specific, with each request consuming more hardware resources (e.g., memory consumption and computation units) due to branch sampling, the number of requests that can be batched together during the decoding phase is much smaller. 
Consequently, the queuing latency bumps for most requests, degrading the overall efficiency.

\section{Motivation and Analysis}
\label{sec:motivation}

In this section, we introduce the observations that motivate our work to address the two challenges discussed in Section~\ref{sec:intro}.

\begin{figure}[!t]
\centering
\includegraphics[width=1.0\textwidth]{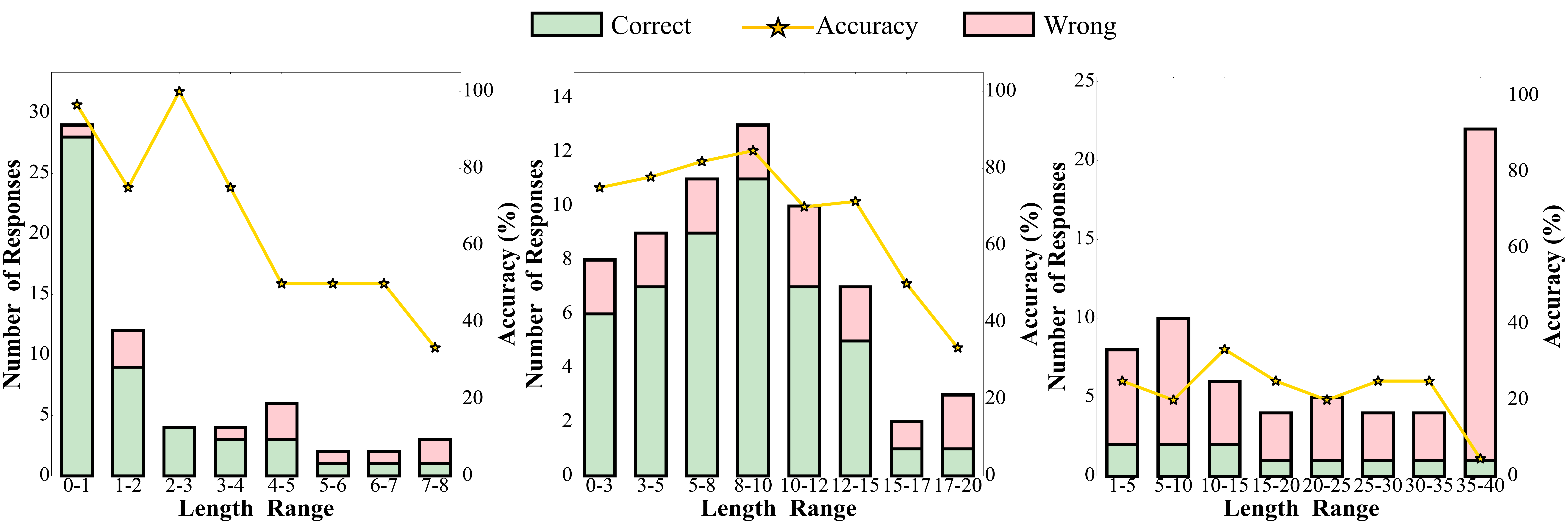}
\caption{\small{
The numbers of correct and wrong responses in each length range, testing with three questions respectively. 
Each $i$-$j$ on $x$-axis denotes a range of response lengths (e.g., ``7-8'' indicates 7K-8K tokens). 
}}
\label{fig:pdf_lenth}
\end{figure}

\textbf{Observation 1: Weak correlation between response length and quality.}
To begin with, we first investigate the first challenge that the inference latency becomes prolonged due to the long-reasoning branches. 
Specifically, we conduct a small testbed by generating 64 responses given three requests and examining the length and correctness. 
As shown in Figure~\ref{fig:pdf_lenth}, for the same request, different trials of reasoning vary in length substantially. 
However, although generating longer responses facilitates deeper reasoning of LLMs, we empirically observe that the response quality is weakly related to the response length. 
To be specific, the portion of correct responses is irrelevant to the lengths.

\textbf{Solution 1: Redundant sampling with early stopping to avoid excessively long responses.}
Although different branches are usually batched together to perform decoding simultaneously, the time required to generate the final answer is still dependent on the longest one. 
Given the substantially varied response lengths, it would be inefficient to wait for all branches. 
Fortunately, as aforementioned, the correctness of a response is somehow irrelevant to its length. 
In fact, many recent studies have pointed out that reasoning-capable LLMs would occasionally suffer from the over-thinking dilemma, which suggests that excessively long responses may even show worse quality~\cite{yi2025shorterbetter,su2025between,chen2025towards,sui2025stop}. 
Therefore, an intuitive idea is to increase the likelihood of obtaining \textit{short-thinking} branches so that we can get rid of the \textit{long-thinking} ones.

Motivated by this, we propose to sample more branches and early-stop the sampling process once there are sufficient completed responses. 
Specifically, assume that we require $M$ responses to achieve the final answer, then we can sample $N$ ($N>M$) branches simultaneously, and terminate the remaining ones once $M$ responses complete. 
By doing so, the number of required decoding steps is only related to the length of the $M$-th completed response, rather than the longest one among the $N$ branches.
To be formal, we analyze the response lengths based on the following lemma from the order statistic theory~\cite{david2004order_statistics}.
\begin{lemma}
Let $X_1, X_2, \cdots, X_N$ be random variables sampled from a cumulative distribution of $F_X(x)$. The $M$-th smallest value, denoted as $X_{(M)}$, has cumulative distribution 
$F_{X_{(M)}}(x;N) = \sum_{i=M}^N {N \choose i} [F_X(x)]^i[1 - F_X(x)]^{N-i}$.
\end{lemma}
Given an arbitrary request, if we consider the length of one response (branch) conforms to some cumulative distribution, then by employing the approach of redundant sampling with early stopping, the probability of requiring $L$ decoding steps to complete $M$ response over $N$ branches is $F_{X_{(M)}}(L;N)$, which is increasing w.r.t. to $N$.
Consequently, by sampling more branches (i.e., by increasing $N$) in parallel, we can complete $M$ responses with fewer decoding steps, thereby improving the efficiency of LLM reasoning.

\begin{figure}[!t]
\centering
\begin{minipage}{0.55\linewidth}
\centering
\includegraphics[width=\linewidth]{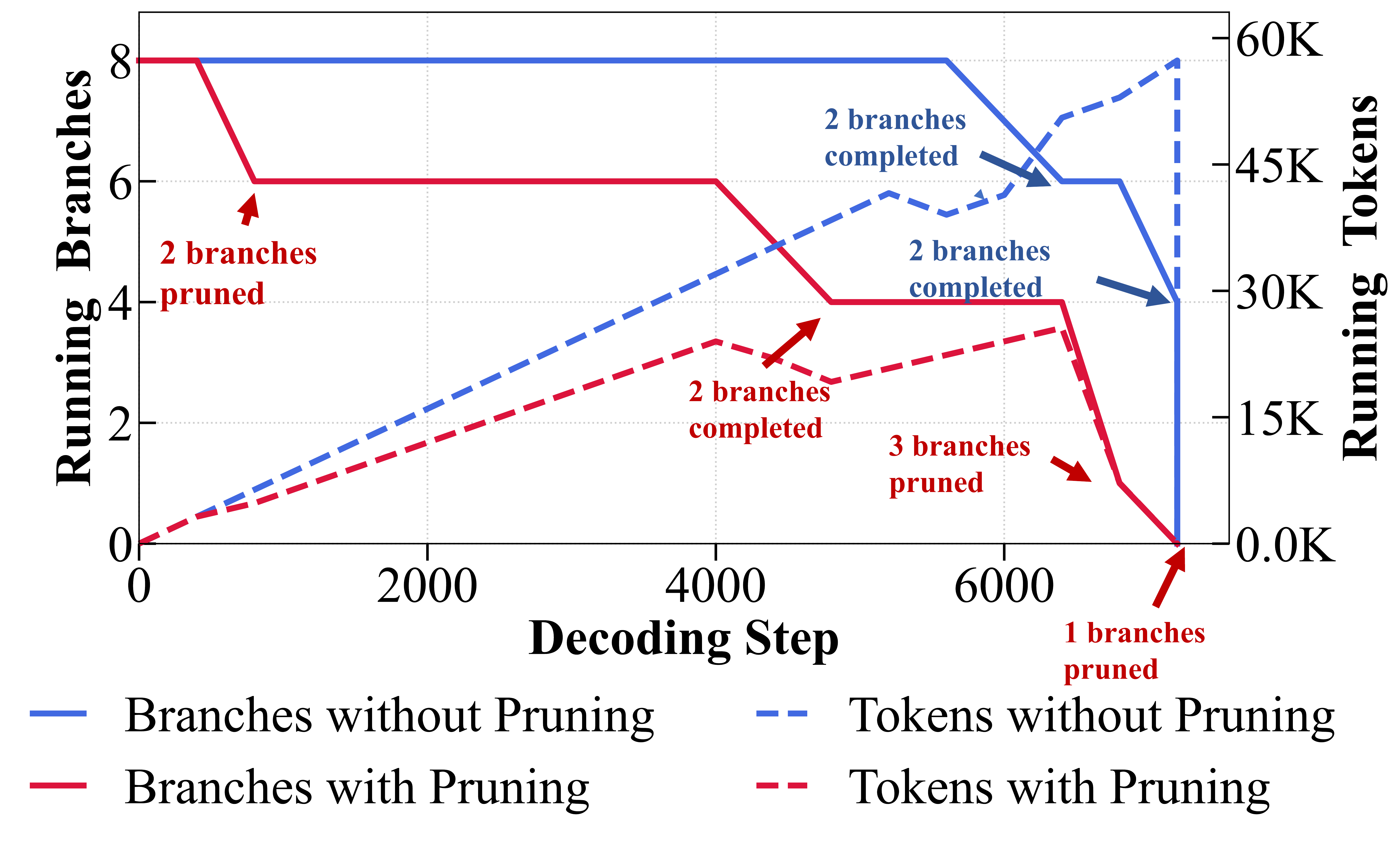}
\caption{\small{The numbers of running branches and tokens with and without pruning. The redundant sampling with early stopping is enabled ($N=8, M=4$).}}
\label{fig:nmprune}
\end{minipage}
\hfill
\begin{minipage}{0.44\linewidth}
\centering
\includegraphics[width=1.0\textwidth]{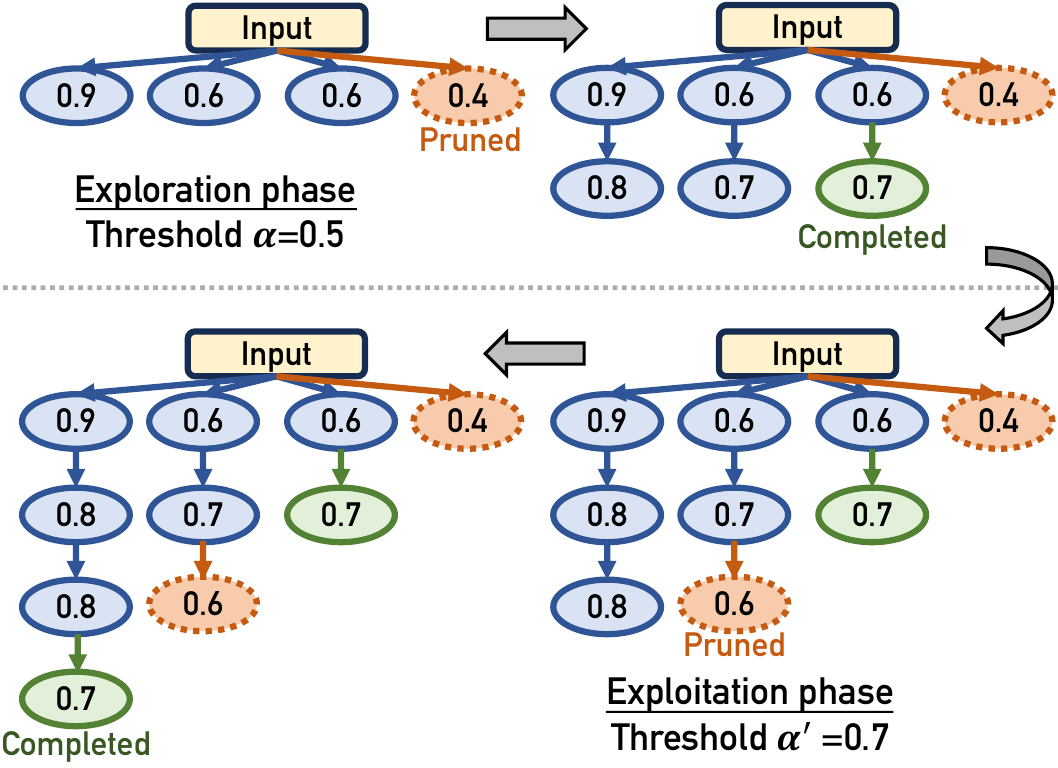}
\caption{\small{An example of our two-phase pruning.
Each value represents the reward of each branch at the corresponding decoding step. 
}}
\label{fig:prune}
\end{minipage}
\end{figure}

\textbf{Observation 2: Extremely low utilization of sampled branches.}
The second challenge is that many requests are forced to queue for a long time since expanding the branches would saturate the hardware resources (e.g., computation and memory). 
Figure~\ref{fig:nmprune} records the number of running branches and tokens to illustrate how the resource consumption of a request changes. 
It can be seen that even though certain branches do not contribute to the final answer (e.g., early-stopped or diverged from the majority), their resource consumption cannot be released until a very late stage.

\textbf{Solution 2: Adaptively pruning low-quality branches for serving efficiency.}
Given such low utilization of sampled branches, there is a natural question that can we prune the branches that are prone to be ineffective during the serving process?
Since the sampled branches would vary in correctness, it essentially requires maintaining \textit{right-thinking} branches and pruning the others so that the final response quality would remain intact. 
Undoubtedly, it is non-trivial to judge whether a reasoning branch is correct or not, yet fortunately, many works have developed Process Reward Models (PRMs) to obtain a reward value given a reasoning process (i.e., a reasoning trajectory). 
Therefore, our work leverages a PRM to measure the quality of each branch.\footnote{It is noteworthy that several existing works have attempted to prune a branch when it undergoes a rethinking process by monitoring whether there are specific tokens/words like ``wait''~\cite{muennighoff2025s1,yang2025dynamic,fu2025reasoning}.
However, such an approach can only give a 0/1 indicator about whether the branch should be pruned, while our work employs PRM since it is able to produce a reward value to facilitate the dynamic change in pruning thresholds.
Besides, using PRM as a predictor can be generalized to broader scenarios compared to monitoring specific tokens/words (e.g., when the LLM is reasoning with another language). As a result, we focus on using PRM in this work.}

Nevertheless, how to prune the branches necessitates careful deliberation. 
In particular, without pruning, our redundant sampling with early stopping approach produces short responses, but with pruning, the response lengths may be higher since a certain amount of short branches get pruned, leading to longer decoding latency from the perspective of each request. 
To balance the decoding latency and queuing delay, we introduce a two-phase dynamic pruning method, which is depicted in Figure~\ref{fig:prune} and detailed below.
\begin{itemize}[leftmargin=*]
\item The first phase prioritizes exploration by setting a low pruning threshold $\alpha$. Only branches with exceptionally low reward values are pruned. 
In addition, to avoid pruning all branches, we set a maximum number of pruned branches $\beta$. 
\item Once a branch has completed, our method moves forward to the second phase by increasing the pruning threshold. Specifically, it updates the threshold as the reward value of the first completed branch (denoted as $\alpha^\prime$). Besides, the constraint of maximum number of pruned branches is ignored in this phase (i.e., equivalent to setting $\beta^\prime = N - 1$), expediting the pruning process. 
\end{itemize}
Note that our pruning method does not guarantee $M$ completed responses. However, we would like to emphasize that this actually fits real-world scenarios since there is no prior knowledge about the difficulty of an incoming request. 
For example, if a short response gets a high reward value, then the request is likely to be easy to answer, so we can skip the long reasoning responses by pruning them in the second phase. 
In contrast, if most branches get extremely low reward values in the first phase, then the request is likely to be hard to solve, as a result of which, our method prunes most branches in the early phase to release the hardware resource consumption and concentrates on the relatively convincing branches even though they are long.

\section{Method}
\label{sec:method}

\begin{algorithm}[!t]
\caption{\small{The scheduling workflow of \name}}
\label{alg:schduler}
\small
\begin{algorithmic}[1]
\Require 
\Statex $N$: the number of sampling branches; $M$: the number of completed branches that triggers early stopping
\Statex $\alpha$: the pruning threshold in the first phase; $\beta$: the maximum number of pruned branches allowed
\Statex $B$: the batch size for decoding; $T$: the number of continuous decoding steps 
\Statex $\text{request\_queue}$: the queue of awaiting requests (inserted by a background thread)
\Statex $\text{branch\_queue}$: the queue of awaiting branches (initialized as empty)
\Statex $\text{meta}$: the dictionary that records the metadata of all processing requests (initialized as empty)

\smallskip

\Statex $\triangleright$ \ul{\textit{Main scheduling workflow}}
\State $\text{current\_branch} \gets []$
\While{not terminated}
\While{${len}(\text{current\_batch}) < B$}
\If{${len}(\text{branch\_queue}) > 0$}
\State $\text{current\_batch}.append(\text{branch\_queue}.pop())$ \Comment{Fill with an awaiting branch}
\ElsIf{${len}(\text{request\_queue}) > 0$}
\State \textsc{\texttt{Prefill}}$(\text{request\_queue}.pop())$ \Comment{Get new branches by prefilling with an awaiting request}
\Else
\State break \Comment{Continue with a smaller batch size since there are no awaiting branches or requests}
\EndIf
\EndWhile
\State \textsc{\texttt{Decode}}$(\text{current\_batch}, T)$
\EndWhile 

\smallskip

\Function{\texttt{Prefill}}{request $i$}
\State Perform prefilling with request $i$
\State $\text{meta}[i] \gets \{
\text{phase=}\text{explore}, 
\text{threshold=}\alpha, 
\text{max\_num\_pruned=}\beta, 
\text{num\_completed=0},
\text{num\_pruned=0}
\}$
\For{$j \gets 1, 2, \cdots, N$}
\State $\text{branch\_queue}.push(b_{ij})$ \Comment{Add $M$ branches for decoding}
\EndFor
\EndFunction

\smallskip

\Function{\texttt{Decode}}{$\text{current\_batch}$}
\State Batch decoding for up to $T$ steps with $\text{current\_batch}$
\For{$i \gets $ involved request indices in $\text{current\_batch}$}
\If{$\text{meta}[i][\text{phase}] == \text{explore}$ \textbf{and} request $i$ has completed branch(es) in $\text{current\_batch}$}
\State Calculate new threshold $\alpha^\prime$ for the second phase
\State $\text{meta}[i].update(
\text{phase=}\text{exploitation}, 
\text{threshold=}\alpha^\prime,
\text{max\_num\_pruned=}N-1
)$ 
\EndIf
\For{$b_{ij} \gets $ completed branch(es)}
\State Remove $b_{ij}$ from $\text{current\_batch}$ \Comment{Remove the completed branch}
\State $\text{meta}[i][\text{num\_completed}] \gets \text{meta}[i][\text{num\_completed}] + 1$
\EndFor
\For{$b_{ij} \gets $ incompleted branch(es)}
\If{$\text{meta}[i][\text{num\_pruned}] < \text{meta}[i][\text{max\_num\_pruned}]$ \textbf{and} $\text{PRM}(b_{ij}) < \text{meta}[i][\text{threshold}]$}
\State Remove $b_{ij}$ from $\text{current\_batch}$ \Comment{Prune the branch}
\State $\text{meta}[i][\text{num\_pruned}] \gets \text{meta}[i][\text{num\_pruned}] + 1$
\EndIf
\EndFor
\If{$\text{meta}[i][\text{num\_completed}] \geq M$ \textbf{or} $\text{meta}[i][\text{num\_completed}] + \text{meta}[i][\text{num\_pruned}] == N$}
\State \textbf{Output} the final response for request $i$ \Comment{$M$ branches completed or no branches remained}
\EndIf
\EndFor
\EndFunction
\end{algorithmic}
\end{algorithm}

Based on the above analysis, we develop a serving framework for efficient and accurate LLM reasoning, namely \name, which employs the redundant sampling with early stopping approach to facilitate obtaining \textit{short-thinking} branches as well as the two-phase dynamic pruning method to prioritize \textit{right-thinking} branches.

The scheduling workflow of \name is shown in Algorithm~\ref{alg:schduler}. 
We treat each branch as one unit for batch decoding. 
When the current batch has a size smaller than a pre-configured batch size $B$, we first try to fill it with awaiting branches (lines 4-5). 
If the number of awaiting branches is not enough to fill the batch, we perform the prefilling process with an awaiting request to obtain $N$ new branches for decoding (lines 6-7). 
An exceptional case is that when there are no awaiting branches nor requests, we will continue the decoding process with a smaller batch size (lines 8-9).

The prefilling process is the same as vanilla LLM inference. 
To facilitate our branch sampling and pruning, we record the metadata for each request, including the current pruning phase, the current pruning threshold, and the numbers of completed and pruned branches (line 16). 
Meanwhile, we append $N$ branches to the queue, which will be scheduled for decoding later (lines 17-19).

To reduce the overhead of reward calculation and pruning, we carry out such routines every $T$ steps. 
The current batch of branches will first undergo up to $T$ decoding steps (line 22).\footnote{We say ``up to'' since some branches may complete with fewer steps.}
Subsequently, we handle each involved request individually to perform pruning or early stopping (line 23). 
To be specific, if the request is current in the first phase for pruning (i.e., the exploration phase), we examine whether it has any branches completed during the last decoding steps (line 24).
If yes, we calculate the reward value of the completed branch(es) and update the pruning phase as well as the pruning threshold (lines 25-26). 
Then, we collect completed branches (lines 28-31), prune low-quality branches if necessary (lines 32-37), and examine whether we can finalize the output for this request due to early stopping (i.e., $M$ branches completed) or pruning (i.e., no branches remained) (lines 38-40).
Last but not least, to optimize KV cache utilization, we share prefix KV cache across branches, and release the corresponding KV cache immediately when a branch is pruned, early-stopped, or completed (the shared prefix KV cache is released only if all branches have terminated).

It is worthy to note that after integrated with continuous batching, the branches of the same request may not start decoding at the same time, which indicates that the earliest completed branch is not necessarily the shortest one. 
However, our redundant sampling with early stopping approach is still applicable since our goal is to sample $M$ completed branches as soon as possible, rather than as short as possible. 
As we will empirically show in our experiments, thanks to our dynamic pruning method, the queuing latency can be substantially reduced, leading to better overall efficiency.

\section{Evaluation}
\label{sec:expr}

\subsection{Implementation and Experimental Setup}
\label{sec:expr_setup}
We implement \name atop vLLM~\cite{kwon2023efficient} and we use Qwen2.5-Math-PRM-7B~\cite{zhang2025lessons} as the PRM for reward calculation. 
After all branches have completed or been pruned, \name selects the one with the highest reward value as the final response. 
By default, we set $M = N / 2$, $\alpha = 0.5$, $\beta = N / 2$, $T = 400$, while $B$ is configured via the workload information (e.g., hardware, model size, etc.). 

\textbf{Hardware environment.}
All experiments are conducted on a server equipped with 8 NVIDIA H100 (80GB) GPUs. The GPUs are interconnected by NVLink with a bandwidth of 900GB/s.

\textbf{Models and workloads.}
We choose two reasoning-capable LLMs, DeepSeek-R1-Distill-Qwen-14B and DeepSeek-R1-Distill-Llama-70B~\cite{guo2025deepseek}, to carry out the experiments. 
Two challenging STEM datasets, GPQA~\cite{rein2023gpqagraduatelevelgoogleproofqa} and GAOKAO~\cite{zhang2024evaluatingperformancelargelanguage}, are used in our evaluation, since they require strong reasoning capabilities to solve. 
For each dataset, we experiment with two different request arrival rates, 1 and 4 requests per second, to represent diverse serving scenarios. 

\textbf{Baselines.}
We mainly compare \name with three baselines: 
(1) Vanilla, which does not involve branch sampling (i.e., $N=1$); 
(2) Self-Consistency~\cite{wang2022self}, which samples $N$ branches in parallel and makes the decision after $N$ responses have been completed; 
and (3) Rebase~\cite{wu2024inference}, which explores the reasoning via tree-based searching with a budget of keeping at most $N$ tree leaves. 
To achieve a fair comparison, we integrate each baseline with continuous batching. 
Specifically, for Self-Consistency and Rebase, we release a branch immediately when it completes (rather than after all branches have completed) to facilitate batching, which is the same as \name.

\textbf{Metrics.}
We focus on both efficiency and accuracy in our experiments. 
For efficiency, we compare the methods by the percentile latencies. 
For instance, P97 latency corresponds to the maximum latency within which 97\% of the requests are served. 
Meanwhile, we measure each method's effectiveness by calculating the accuracy (i.e., the ratio of correctly answered requests). 
We run five trials for each experiment and report the mean (all standard deviations are within 10\%).

\begin{figure}[!t]
\centering
\includegraphics[width=1.0\textwidth]{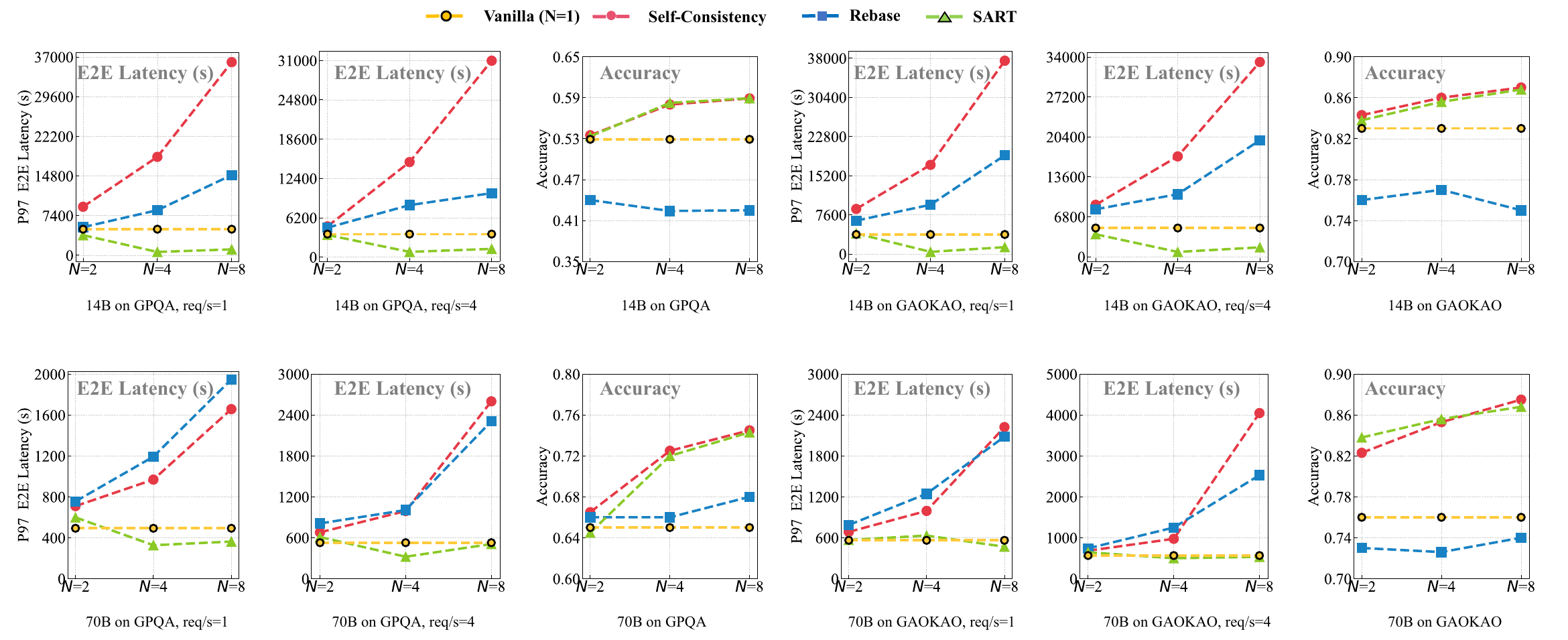}
\caption{\small{
End-to-end (E2E) latency (lower is better) and accuracy (higher is better) of each method with different $N$. 
We plot a horizontal line for Vanilla (which corresponds to $N=1$) to indicate the baseline performance of reasoning without branch sampling.
}}
\label{fig:expr_e2e}
\end{figure}

\subsection{End-to-end Comparison}
\label{sec:expr_e2e}

We first assess the efficiency and accuracy of all methods by tuning $N$. 
The results are shown in Figure~\ref{fig:expr_e2e}. 
In general, \name consistently achieves both better efficiency and higher accuracy compared to Vanilla, demonstrating its superior performance.

When $N$ increases, Self-Consistency and Rebase typically exhibit lowering efficiency due to the heavier workload of branch sampling. 
In contrast, the end-to-end latency of \name is robust to the value of $N$, and in most cases, is lower than that of Vanilla. 
This is not surprising since \name is able to produce short responses and decrease the queuing time well (detailed shown in Section~\ref{sec:expr_ablation}). 
Eventually, when $N=8$, \name outperforms Vanilla, Self-Consistency, and Rebase by up to 3.1$\times$, 28.2$\times$, and 14.4$\times$ and on average 2.0$\times$, 15.7$\times$, and 8.0$\times$, respectively. 

In terms of accuracy, both \name and Self-Consistency outperform Vanilla and achieve higher accuracy as $N$ increases, which is reasonable as a larger $N$ often leads to a larger amount of test-time computation. 
Although \name gives slightly lower accuracy than Self-Consistency in several cases, the gap is very minor (within 1.6\%). 
Besides, recalling the significant speedup achieved by \name, we believe the performance of our work is still preferable. 
Last but not least, the scaling of Rebase is undesirable and its accuracy is even lower than Vanilla in many cases. 
This may be due to the fact that longer responses would exponentially enlarge the search space of Rebase, making its searching approach less effective. 
(To elaborate, the responses reported in~\cite{wu2024inference} only consist of dozens or hundreds of tokens, while those in our experiments usually contain thousands of tokens given that state-of-the-art LLMs are getting more reasoning-capable.)

\begin{figure}[!t]
\centering
\includegraphics[width=1.0\textwidth]{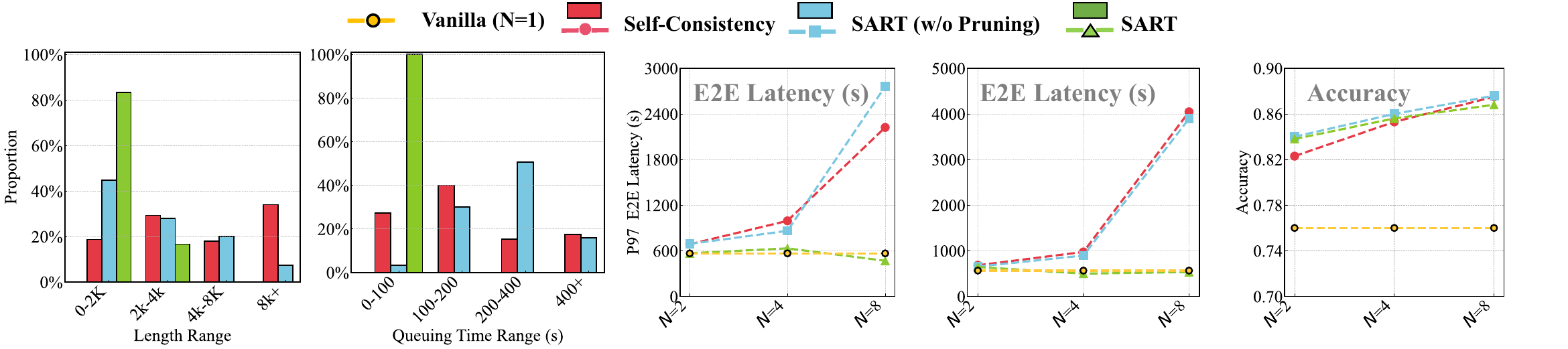}
\caption{\small{Ablation studies (the 70B model on GAOKAO).
The two plots on the left present the distributions of response length and queuing time, respectively ($N=4$ for Self-Consistency, $N=8, M=4$ for \name). 
The three plots on the right present the end-to-end (E2E) latency and accuracy with different $N$.
``\name (w/o Pruning)'' indicates a variant of \name that enables the redundant sampling with early stopping but disables the two-phase dynamic pruning.}}
\label{fig:compare}
\end{figure}

\subsection{Ablation Studies}
\label{sec:expr_ablation}

Next, we conduct ablation studies to assess the effectiveness of the proposed techniques using the GAOKAO dataset and the 70B model. 
As shown in Figure~\ref{fig:compare}, we present a case study to investigate the distributions of response length and queuing time (the two plots on the left), and compare the end-to-end latency and accuracy of all methods (the three plots on the right).

It can be seen that our approach of redundant sampling with early stopping effectively reduces the overall response lengths compared to Self-Consistency, which matches the analysis in Section~\ref{sec:motivation}. 
Besides, the accuracy is comparable to Self-Consistency, which verifies that prioritizing short responses would not affect the final response quality. 
However, it increases the queuing time since more branches are sampled, and therefore, it only achieves similar efficiency as Self-Consistency. 
Fortunately, our two-phase dynamic pruning method comes to help by shrinking the queuing time, thereby reducing the end-to-end latency greatly. 
More importantly, the accuracy remains stable even though we allow branches to be pruned, demonstrating that eliminating low-quality branches does not harm the final response quality.

\begin{figure}[!t]
\centering
\includegraphics[width=1.0\textwidth]{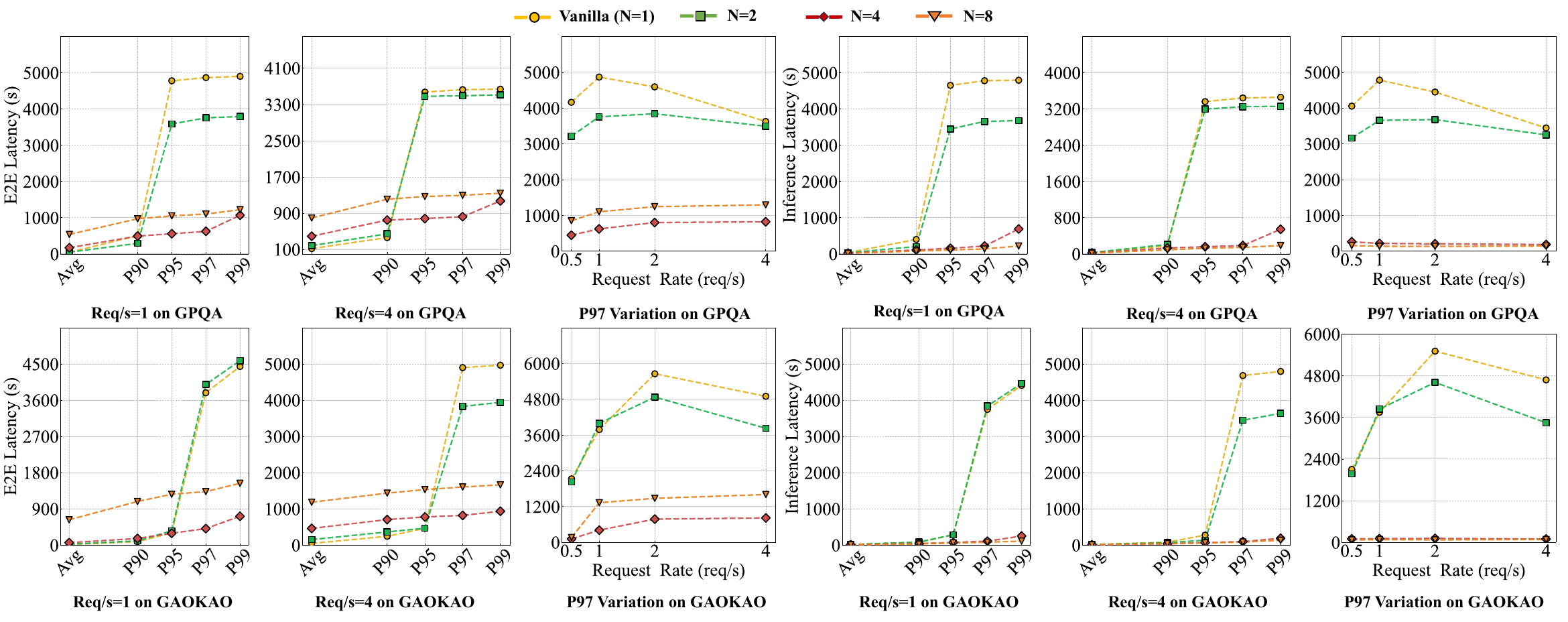}
\caption{\small{
End-to-end (E2E) latency and inference latency (i.e., E2E latency excluding queuing delay) under various $N$, evaluated with the 14B model.
}}
\label{fig:sensitiveplot}
\end{figure}

\subsection{Sensitivity Analysis}
\label{sec:expr_sensitivity}

Finally, we evaluate the sensitivity of our work w.r.t. the number the branches $N$. 
As shown in Figure~\ref{fig:sensitiveplot}, when $N\in\{4,8\}$, the average (i.e., P50) and P90 latencies are higher than those of $N\in\{1,2\}$ as more computation FLOPs are needed. 
However, the tail latencies (P97 and P99 latencies) when $N\in\{4,8\}$ are much lower, which is reasonable since \name effectively avoids excessively long responses and reduces queuing time. 
Last but not least, although $N=8$ achieves lower inference latencies (i.e., the latency excluding queuing) than $N=4$, the end-to-end latency (including queuing) is higher. 
This shows that although a large $N$ helps the model find high-quality responses within shorter lengths, the increase in queuing delay would degrade the overall efficiency. 
Nevertheless, in general, the performance gap between $N=4$ and $N=8$ is small, demonstrating robustness w.r.t. the choice of $N$.

\section{Conclusion and Limitations}
\label{sec:conc}

In this work, we introduced \name, a serving framework for efficient and accurate LLM reasoning. 
Following empirical observations and analysis, we proposed a redundant sampling with early stopping approach to increase the likelihood of sampling short-thinking branches, as well as a two-phase dynamic pruning method to prioritize sampling right-thinking branches. 
Empirical results show that \name achieves up to 28.2$\times$ and on average 15.7$\times$ of speedup when achieving the same level of accuracy compared to existing works. 

Despite its effectiveness, \name still has several limitations. 
Firstly, it introduces additional hyper-parameters, which may increase the complexity of performance tuning. 
Secondly, it processes the requests in a first-come-first-served manner, and requires redesigns if we need to support preemptible scheduling strategies. 
We would like to explore and address these issues in the future.

\bibliographystyle{plainnat} 
\bibliography{cite} 

\end{document}